\documentclass[letterpaper, 10 pt, conference]{ieeeconf}  

\IEEEoverridecommandlockouts                              
\usepackage[utf8]{inputenc}
\usepackage[T1]{fontenc}
\usepackage{verbatim}
\usepackage{graphicx} 
\usepackage{amssymb}  
\usepackage{hyperref}
\usepackage{xcolor}
\usepackage[export]{adjustbox}
\newcommand{\TODO}[1]{}
\renewcommand{\TODO}[1]{{\color{red}TODO: {#1}}}
\newcommand{\WC}[1]{}
\renewcommand{\WC}[1]{{\color{purple} {#1} (wc?)}}

\title{\LARGE \bf
Outlier detection of vital sign trajectories from COVID-19 patients
}

\author{Sara Summerton$^{1}$, Ann Tivey$^{2}$, Rohan Shotton$^{2}$, Gavin Brown$^{1}$, Oliver C. Redfern$^{3}$, Rachel Oakley$^{2}$, \\ John Radford$^{2,4}$, and David C. Wong$^{1,5}$
\thanks{$^{1}$S. Summerton, G. Brown and D. C. Wong are with the Department of Computer Science, University of Manchester, Kilburn Building, Oxford Road, Manchester, UK.
	{\tt\small sara.summerton@manchester.ac.uk}}%
\thanks{$^{2}$A. Tivey, R. Shotton, R. Oakley, and J. Radford are with the Christie NHS Foundation Trust, Manchester Academic Health Science Centre, Manchester, UK}%
\thanks{$^{3}$O. Redfern is with the Nuffield Department of Clinical Neurosciences, University of Oxford, Kadoorie Centre, John Radcliffe Hospital, Oxford, UK.}%
\thanks{$^{4}$J. Radford is also with the Institute of Cancer Sciences, University of Manchester, Manchester, UK.}%
\thanks{$^{4}$D. C. Wong is also with the Centre for Health Informatics, University of Manchester, Vaughan House, Portsmouth Street, Manchester, UK.}%
}

\begin{document}

\maketitle
\thispagestyle{empty}
\pagestyle{empty}

\begin{abstract}
	In this work, we present a novel trajectory comparison algorithm to identify abnormal vital sign trends, with the aim of improving recognition of deteriorating health. 
	
    There is growing interest in continuous wearable vital sign sensors for monitoring patients remotely at home. These monitors are usually coupled to an alerting system, which is triggered when vital sign measurements fall outside a predefined normal range. 
	Trends in vital signs, such as increasing heart rate, are often indicative of deteriorating health, but are rarely incorporated into alerting systems. 
	
	We introduce a dynamic time warp distance-based measure to compare time series trajectories. We split each multi-variable sign time series into 180 minute, non-overlapping epochs. We then calculate the distance between all pairs of epochs. Each epoch is characterized by its mean pairwise distance (average link distance) to all other epochs, with clusters forming with nearby epochs.
	
	We demonstrate in  synthetically generated data that this method can identify abnormal epochs and cluster epochs with similar trajectories. We then apply this method to a real-world data set of vital signs from 8 patients who had recently been discharged from hospital after contracting COVID-19. We show how outlier epochs correspond well with the abnormal vital signs and identify patients who were subsequently readmitted to hospital.

\end{abstract}

\section{INTRODUCTION}

Monitoring vital signs is commonly used in clinical practice to aid assessment of a patient's condition. 
Abnormal vital signs are known to precede adverse events. \cite{schein1990clinical}. For instance, in patients with COVID-19, there is strong correlation between decreasing oxygen saturations (SpO2) and severe cases that require hospitalization \cite{greenhalgh2021remote}.

Traditionally, detection of abnormal vital signs is assessed using Early Warning Score (EWSs) calculated from intermittent, manually-collected measurements \cite{downey2017strengths}. 
One key limitation of EWSs is that they are typically calculated from only the most recent set of vital signs \cite{gerry2020early}. This has traditionally been the case even when EWSs are incorporated into continuous monitoring devices \cite{tarassenko2006integrated}. It is possible that trends in vital signs may enable earlier detection of deterioration, and yield insights into different patterns of deterioration. Previous studies of hospitalized patients have attempted to include information about vital sign trends. However, these have used relatively simplistic summaries, such as the difference of current and baseline values \cite{charbonnier2010line}\cite{bell2021trend}. 

In this exploratory analysis, we assess how trends in vital signs over time (trajectories), regardless of the absolute value, could identify and distinguish between different patterns of patient deterioration. We propose an approach to outlier detection for vital sign trajectories and evaluate its performance on synthetically generated multivariate time series. We then consider its behavior when applied to continuous vital sign data collected from COVID-19 patients that had been discharged from hospital to their own home. We demonstrate that this method correctly identifies outliers and clusters trajectories on both data sets, illustrating the potential of this method to identify COVID-19 deterioration.

\section{METHODS}
In the following section we first introduce our overall approach for identifying outlier trends.
We created synthetic vital sign data with known abnormalities to study the behavior of this method. We describe the generation process later in this section.
We then describe the clinical vital sign data, as well as preprocessing steps taken, in more detail.

\subsection{Outlier Detection}
We implement an outlier detection approach based on distance to nearest neighbors. This family of approaches is described in detail by Pimentel et al. \cite{pimentel2014review}. Consider a sequence of data $A = [a_1, a_2, ..., a_m]$, in which $a_i \in \mathbb{R}^n$ is an $n$-dimensional feature vector. We denote the similarity between two such sequences, $A$ and $B$, as some function $F(A, B)$. One way to define similarity is as a distance such as the Dynamic Time Warp (DTW) distance, 

\[F(A, B) = \sqrt{ \sum_{(i,j)\in \pi} ||a_i - b_j||^2} \, , \]

\noindent with $1\leq i,j \leq m$, and where $\pi$ is the optimal alignment path, defined as the contiguous path through the matrix of squared element-wise differences between both sequences that minimizes the cumulative distance between them \cite{Ratanamahatana2004}. The advantage of DTW over the Euclidean distance is that it allows nonlinear alignments, so similar but non-aligned or out of phase sequences can be meaningfully compared. The DTW distance can accommodate time series of unequal length, however all sequences in this work were fixed to the same length. 

A hierarchical clustering approach can be taken to assess the similarity of a fixed length multivariate time series in the context of a set of multiple time series. Agglomerative clustering calculates the distance between each time series, then joins the pair of time series with the shortest distance into a single cluster in an iterative process until the entire dataset is contained in a single cluster. The distance between two clusters is defined as $$ D(U, V) = \frac{1}{|U|\cdot|V|} \sum_{u\in U}\sum_{u \in V} F(u,v) $$ where $u$ and $v$ are elements and $|U|$ and $|V|$ are the cardinalities of clusters $U$ and $V$, respectively, and $D$ is the average-linkage distance. $U$ and $V$ can be a cluster of multiple time series or a single sequence. A stopping criteria can be provided by setting a maximum accepted value of $D$ for agglomerating new clusters.  

Many real-world clinical problems involve detecting abnormal physiological signals in an abundance of normal data. Patients with stable vital signs will comprise the majority of the time series segments, and we expect these series to have low average-linkage distances and thus be clustered together first. The time series most dissimilar to the cluster of normal data will have the largest average-linkage distance, and it is these final time series or clusters that may implicate outliers in the data corresponding to abnormal physiological signals.

\subsection{Data generation, collection, and analysis}

\begin{table}\begin{center}
		\caption{Perturbations added to synthetic HR (blue) and RR (orange)}
		\label{tab:synth}
		\begin{tabular}{|c|c|}
			\hline \rule{0pt}{2.3ex}
			Type &  Shape (without noise) \\
			\hline \rule{0pt}{4.3ex}%
			1. Step function (HR) & \raisebox{-0.76\totalheight}[0pt][0pt]{%
				\includegraphics[width=0.2\textwidth, height=21mm]{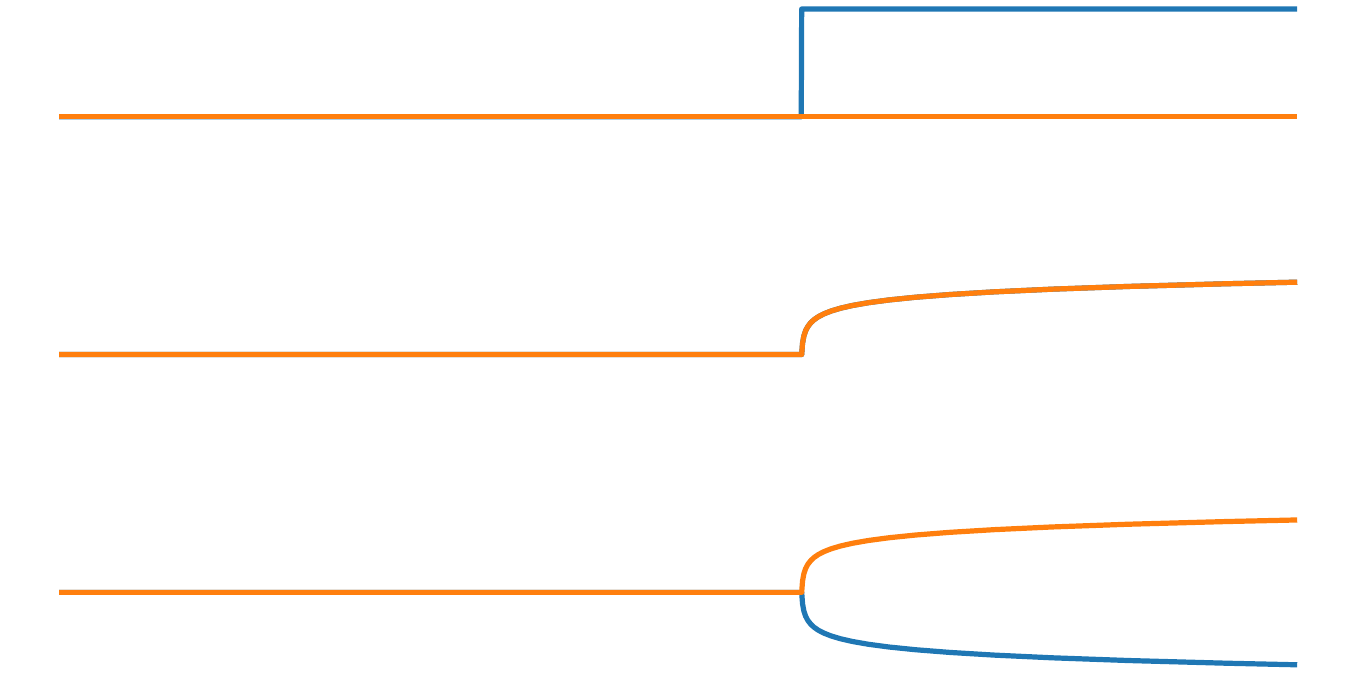}} \rule[-1.5ex]{0pt}{-3ex} \\ 
			\hline \rule{0pt}{4.3ex}%
			2. Increase HR, RR & \rule[-1.5ex]{0pt}{-3ex} \\ 
			\hline \rule{0pt}{4.3ex}%
			3. Increase HR, decrease RR & \rule[-1.6ex]{0pt}{-3ex} \\ 
			\hline
		\end{tabular}
		\vspace{-4mm}
	\end{center}
\end{table}

\textit{Synthetic Data} --- Synthetic heart rate (HR) and respiration rate (RR) data were generated for 20 `patients' from a mixture of two periodic functions with Gaussian noise added. The modulating function had a period of one day, to which a second, low amplitude sine wave was added with a period of four hours. The noise was sampled from a 2D Gaussian with unit variance and random (positive, symmetric) covariance, restricted to the range [0.15, 0.6] so as to adhere to a similar covariance as the wearable sensors data after normalization. 

Each synthetic data file has a duration of 8 days (3840 hours). Methods of perturbation and values to which they were applied are listed in Table \ref{tab:synth}. Perturbations were added to the last 10\% (approx. 19 hours) of six files, two for each type of perturbation, for a total of 115 hours of `abnormal' data. The magnitude of each perturbation was scaled to approximately 1.5 standard deviations of the original distribution.  

\textit{Wearable vital sign pilot data} --- Vital sign data were collected from cancer patients who had contracted COVID-19, been admitted to hospital, and had been subsequently considered suitable for outpatient care. All participants wore Isansys\textsuperscript{TM} sensors which recorded their heart rate (HR), respiratory rate (RR) and temperature (Temp) each minute for up to three weeks. 

In total, data were recorded from eight patients as part of the RECAP pilot study. The study is listed on the ClinicalTrials.gov registry with study ID NCT043977052 (\url{https://clinicaltrials.gov/ct2/show/NCT04397705}). All study participants provided signed written consent.

\textit{Data Processing and Analysis} --- The same pre-processing was carried out on both the synthetic and the real data. First, each channel was normalized on a per-patient basis to zero mean and unit variance, then low-pass filtered using a 25-point (i.e. 25 minute) median filter to remove short-term fluctuations in heart and respiration rate, likely caused by movement artefacts and sensor noise. Second, the signals were segmented into 180-minute epochs. Data segments that were shorter than the epoch length were discarded. The epoch length was chosen so as not to capture medium-term variations in vital sign data, such as transitory increases in heart rate due to short-term physical activity, but still encapsulate overall changes in physiological condition.

The described outlier detection approach was then applied to all epochs, ranking each epoch by hierarchical average-linkage distance using the DTW distance measure. The maximum intra-cluster distance, and thus the number of clusters, was chosen based on the step in agglomerative clustering associated with the maximum difference in average-linkage distance \cite{Zambelli2016}.

To visualize similarity between epochs, we used multidimensional scaling (MDS), which is a dimensionality reduction approach that seeks to preserve the distance between data in the original high-dimensional space, in this case, the matrix of DTW distances between epochs \cite{kruskal1978multidimensional}. We used this to describe and examine the sequence of contiguous epochs for representative patients from the dataset.


All data generation and processing was undertaken in Python using the \textit{scipy}, \textit{sklearn}, and \textit{tslearn} libraries. Code supporting this article is available at \url{https://github.com/sara-es/outlier-detection-RECAP-data}.

\begin{figure*}[h!]
	\centering
	\includegraphics[scale=0.52, center]{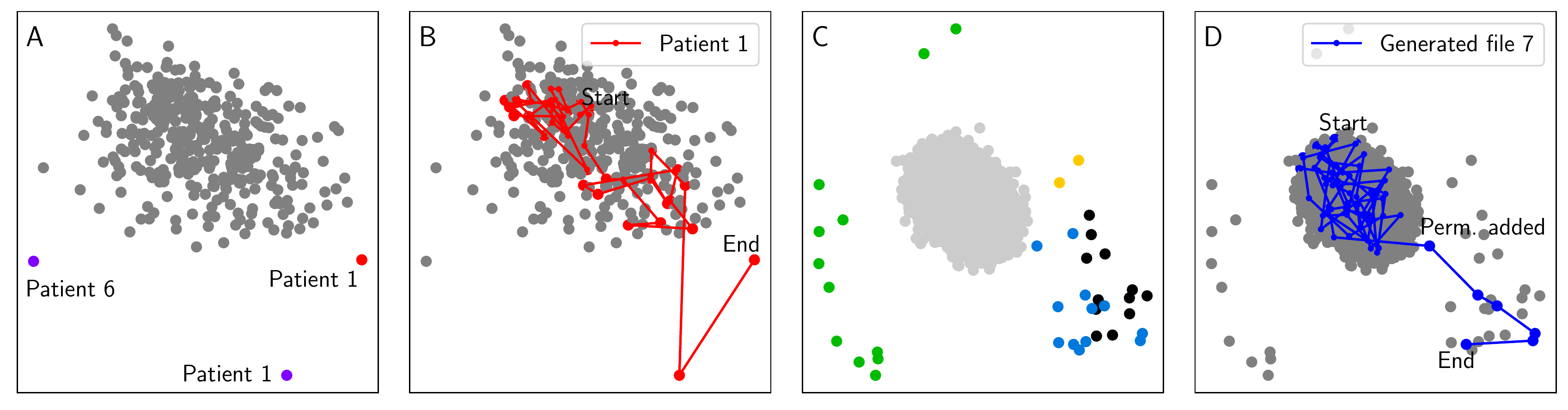}
    \vspace{-6mm}
	\caption{Distances between epochs are visualized in 2D via MDS maps, on which an individual point represents a 180-minute segment of real (plots A, B) or synthetic (plots C, D) heart rate and respiratory rate data. (A) Epochs from the real data are shown in colors corresponding to their cluster membership; the majority of epochs form a central cluster (shown in gray), and one cluster (shown in red) contains only one epoch. (B) Epochs from patient 1, who was readmitted to hospital, have been highlighted in red and connected in consecutive order. The initial 180-minute epoch is denoted by the text `start', and the final epoch is denoted by the text `end'. The final two epochs of collected data can be shown to rest outside the central cluster of time series epochs. Epochs from all other patients are in gray. (C) Epochs from the synthetic data are shown in colors corresponding to their cluster membership. \emph{Grey}: no permutation (100\% of epochs initialized in this group); \emph{black}: type 1 (83\%); \emph{yellow}: type 1 (17\%); \emph{blue}: type 2 (100\%); \emph{green}: type 3 (100\%). The yellow cluster corresponds to epochs with the step function change point in permutation type 1. (D) Epochs from a single generated file have been highlighted in blue and connected in consecutive order. The epochs to which a permutation was added, near the end of time series, can be shown to progress outside the central cluster of time series epochs.
	}
	\label{fig:mds}
\end{figure*}

\begin{table}\begin{center}
		\caption{Patient-level overview of amount of vital sign data recorded (hours) and clinical events (hospital readmission)}
		\label{tab:recap}
		\begin{tabular}{|c||c|c|c|c|}
			\hline \rule{0pt}{2.3ex}
			ID & HR & RR  & Temp. & Events\\
			\hline \hline \rule{0pt}{2.3ex}
			1 & 146 & 146 & 154 & Hospital\\
			\hline\rule{0pt}{2.3ex}
			2 & 2 & 2 & 2 & None\\
			\hline\rule{0pt}{2.3ex}
			3 & 2 & 2 & 17 & None\\
			\hline\rule{0pt}{2.3ex}
			4 & 300 & 300 & 107 & None\\
			\hline\rule{0pt}{2.3ex}
			5 & 28 & 28 & 41 & Hospital\\
			\hline\rule{0pt}{2.3ex}
			6 & 280 & 280 & 305 & None\\
			\hline\rule{0pt}{2.3ex}
			7 & 369 & 369 & 402 & None\\
			\hline\rule{0pt}{2.3ex}
			8 & 156 & 156 & 257 & None\\ 
			\hline
		\end{tabular}
		\vspace{-8mm} 
	\end{center}
\end{table}

\section{RESULTS}
\subsection{Synthetic data}
We applied our outlier detection approach to the synthetic data, which resulted in five clusters.  Epochs with Type 2 and 3 perturbations, as well as all `normal' epochs, each formed distinct clusters. 
A fourth cluster was composed of 83\% (10/12) of the epochs with a Type 1 perturbation; the remaining two formed their own cluster. We note that the two epochs in this final cluster were the `transitional' epochs between the base function and base function with Type 1 perturbation added. 
Figure \ref{fig:mds}C shows the MDS map of all synthetic data epochs color coded by cluster membership, and Figure \ref{fig:mds}D shows the progression of one time series with a Type 2 perturbation of later epochs. 

\subsection{Wearable vital sign data}
Table \ref{tab:recap} shows the duration of vital sign data recorded for each patient in the data set, as well as whether the patient was readmitted to hospital. In total, there were 1561 patient-hours of data for the 8 patients; the mean length of data recording was 230 hours (range: 2.3 to 527 hours).

\begin{figure*}[!h]
	\centering
	\includegraphics[scale=0.53, center]{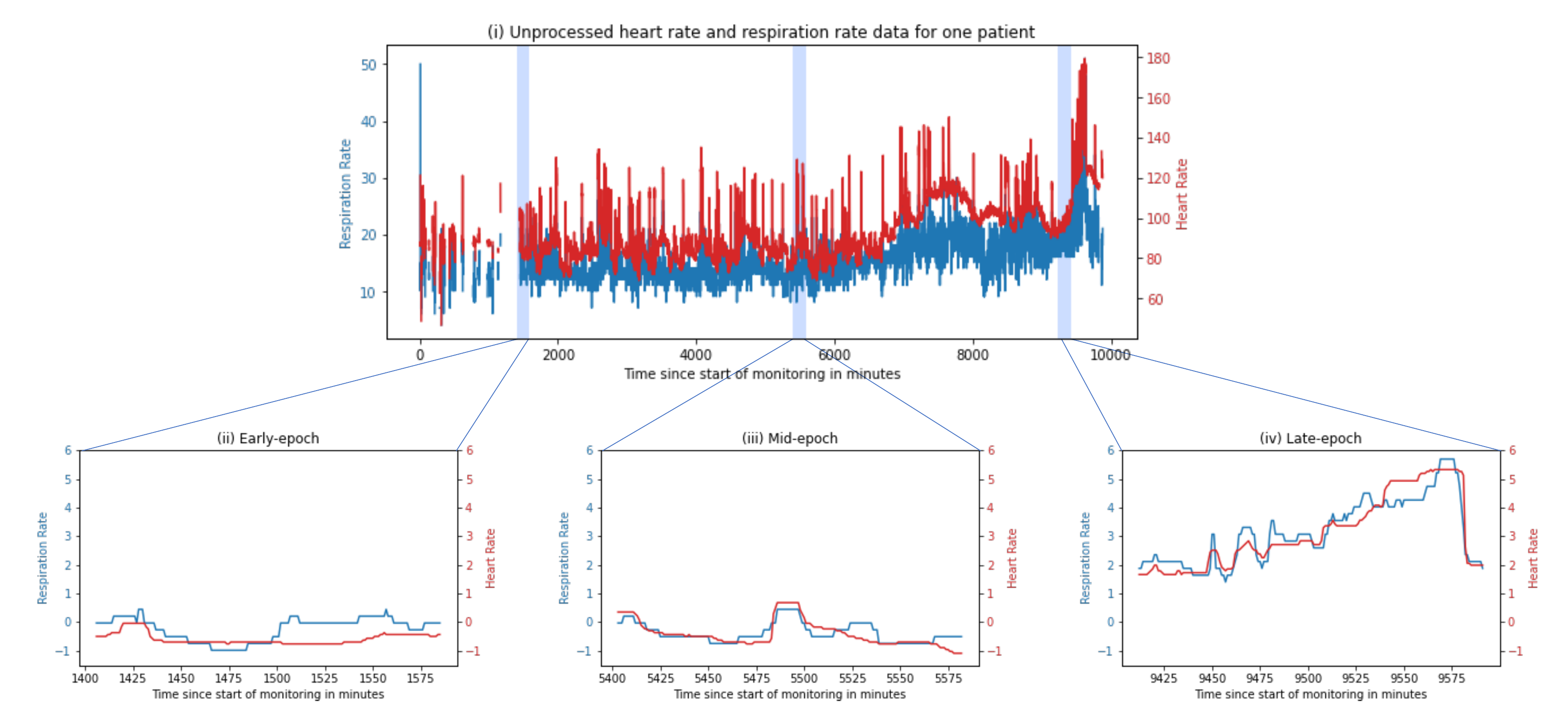}
	\caption{(i) The raw HR and RR data for patient 1 over the entire duration of monitoring (approximately 7 days). (ii) The \emph{start} epoch of normalized and smoothed HR and RR data for patient 1. (iii) An epoch taken from the approximate mid point of monitoring data. (iv) The \emph{end} epoch for patient 1, showing visible deviation from previous baseline measurements. }
	\label{fig:vitals-pt1}
\end{figure*}

We checked the data quality of HR, RR and Temp recordings by plotting their distributions. Based on this, we excluded temperature, as the data contained a high proportion of physiologically implausible values (20.8\% were lower than 34 C). Poor data quality from skin temperature sensors in wearable devices is a known issue \cite{downey2019reliability}.

We applied our outlier detection approach to the HR and RR data, which resulted in three clusters, one containing a single epoch, as shown in Figure \ref{fig:mds}A. Of the three outliers, two belong to patient 1, who was readmitted to hospital. The other is from to Patient 6, who was not readmitted.

Figure \ref{fig:mds}B shows the MDS map of time series epochs from all patients. The sequence of contiguous epochs for Patient 1 has been highlighted in red. Patient 1's initial epoch lies towards the centre of the MDS map, indicating that it is similar to multiple other epochs. Towards the end of the monitoring period, the epochs progress away from the starting location on the map. The final two epochs are far away from all other points on the MDS map, indicating a highly unusual trajectory.  

The raw time series epochs corresponding to the start and end points of patient 1, as well as one intermediate epoch, are shown in Figure  \ref{fig:vitals-pt1}. We observe that the `start' epoch contains HR and RR trajectories that are both relatively flat. In contrast, the `end' epoch contains vital signs that have deviated from their baseline average, and trajectories for both increase across the epoch. This trajectory is visually very different to the start and intermediate epochs in the figure, confirming the validity of the outlier detection approach.

\section{DISCUSSION}
We developed a novel method to identify abnormal multivariate vital sign time series. Unlike previous methods that use categorical variables or change scores to incorporate trends, our method considers the entire shape of a time series epoch via the DTW distance. By clustering based on this distance, we can determine outlying, unusual epochs. Furthermore, this method can group distinct trend patterns, demonstrated by the fact that the different perturbations introduced to the synthetic data resulted in distinct clusters.
We demonstrated the efficacy of this approach on both synthetically generated and real clinical data, successfully identifying abnormal trajectories in both cases. 

When applied to a small patient cohort, this approach yielded promising initial results. Of the 1\% of most outlying epochs, 2/3 belonged to a patient who went on to require hospital admission within 24 hours of the end of monitoring. Furthermore, our per-patient visualization showed how epochs became progressively more abnormal for a patient who required readmission to hospital. These results therefore provide descriptive early evidence suggesting our approach for assessing vital sign trends may be useful for predicting COVID-19 deterioration.

While our approach to detect abnormal vital sign trajectories shows promise, there are several limitations. First, we chose epoch lengths of 180 minutes, based on clinical judgement. However, there is no guarantee that this epoch length is optimum. Second, we used DTW distances to compare epochs, when other distance measures may be more appropriate or faster to compute.
Both the epoch length and distance measure can be optimized via cross-validation. The current data set was insufficient to attempt this, as data comprised only 8 patients and only 2 were readmitted to hospital (positive events).

Finally, we note that this method did not highlight any epochs from the second patient to be readmitted (patient 5). However, only 28 hours of HR and RR data were collected for this patient at the start of the monitoring period. At this point the patient withdrew from data collection due to sensor discomfort and no data were recorded for the three days before their readmission to hospital. 
Therefore, it is likely the decision to readmit was founded on external information not evident in the available HR and RR data.  

In this work, we introduced a trajectory comparison algorithm that is capable of identifying outlying trends in time series. This method distinguishes between different abnormal trajectories, forming multiple and distinct clusters of outliers. By applying this algorithm to real outpatient data, we showed that vital sign trajectories may contain clinically relevant information, predictive of patient deterioration. Future work should apply our method to larger data sets with more positive clinical events.

\addtolength{\textheight}{-12cm}   





\section*{\small{ACKNOWLEDGMENT}}
\vspace{-1mm}
We are grateful for technical support provided by Isansys. We are grateful for the support of the Christie digital services department in facilitating this study. The authors have no conflicts of interest to declare.


\bibliography{IEEEabrv.bib,main.bib}{}
\bibliographystyle{IEEEtran}


\end{document}